# 3D-UGCN: A Unified Graph Convolutional Network for Robust 3D Human Pose Estimation from Monocular RGB Images

Jie Zhao[1], Jianing Li[1], Weihan Chen[1], Wentong Wang[1*], Pengfei Yuan[2], Xu Zhang[2], Deshu Peng[2]

[1]Academy of Artificial Intelligence, Beijing Institute of Petrochemical Technology, Beijing 102617, China
[2] China Year (Beijing) Technology Co.,Ltd., Beijing, China
*Corresponding author: wangwentong@bipt.edu.cn

**Abstract:** Human pose estimation remains a multifaceted challenge in computer vision, pivotal across diverse domains such as behavior recognition, human-computer interaction, and pedestrian tracking. This paper proposes an improved method based on the spatial-temporal graph convolution net-work (UGCN) to address the issue of missing human posture skeleton sequences in single-view videos. We present the improved UGCN, which allows the network to process 3D human pose data and improves the 3D human pose skeleton sequence, thereby resolving the occlusion issue.

## 1 Introduction

Human position estimate is a major difficulty in computer vision. This technology is essential to several areas, including human-computer inter-face, behavior identification, and pedestrian tracking. Virtual live streaming and gaming are two other applications for it. Even in complex situations, Gao W et al.'s[1] spatial-temporal graph convolution model enhances worker safety identification in real-world work scenarios. This provides insightful safety observation data for real-world operations.

Nevertheless, human body position estimation is one of the most difficult jobs in computer vision due to the complexity and variety of human motion, which is frequently executed by stationary individuals[2]. The process of estimating human poses by using pictures, videos, or webcam broadcasts as inputs is known as hu-man pose estimation. There are three types of estimate for three-dimensional human pose: estimation based on 2D information, estimation based on direct regression, and hybrid approaches.

One obvious issue in 3D human posture estimation is occlusion, which arises from several sources of interference or occlusion from different angles. A multi-view 3D posture estimation technique with a restricted number of traditional cameras is presented by Yang et al.[3].

This paper proposes an improved method based on the spatial-temporal graph convolution net-work (UGCN) to address the issue of missing human posture skeleton sequences in single-view videos. This allows the improved network model to achieve perfect 3D human posture estimation. After reviewing the method of 3D human posture estimation based on 2D skeleton sequences, the working principle of spatial-temporal graph convolution is discussed. We present the improved UGCN, which allows the network to process 3D human pose data and improves the 3D human pose skeleton sequence, thereby re-solving the occlusion issue. The following categories best describe the primary contributions of this paper: Firstly, to tackle the problems caused by partial human pose. estimation loss and occlusion in single-view 3D human pose estimation, this study offers a 3D-UGCN model, which is an extension of the input frames and a modification of the input UGCN data into a 3D hu-man skeleton sequence. In order to get more reliable and accurate pose estimation results, the method presented in this research fully utilizes the 3D-UGCN model and the direct regression-based methodology. Second, a 3D-UGCN architecture is suggested in this research that can both effectively increase the accuracy of 3D human pose estimation and refine the 3D human skeleton sequence. Lastly, a plethora of trials and comparisons with other ways show how effective the suggested strategy is in this study.

## 2 Related work

### 2.1 2d-3d attitude estimation

It is possible to categorize 3D human pose estimation based on 2D information into pose estimation networks based on 2D skeleton sequence input combined with 3D and 2D information. The pose estimation network integrates 2D and 3D information.

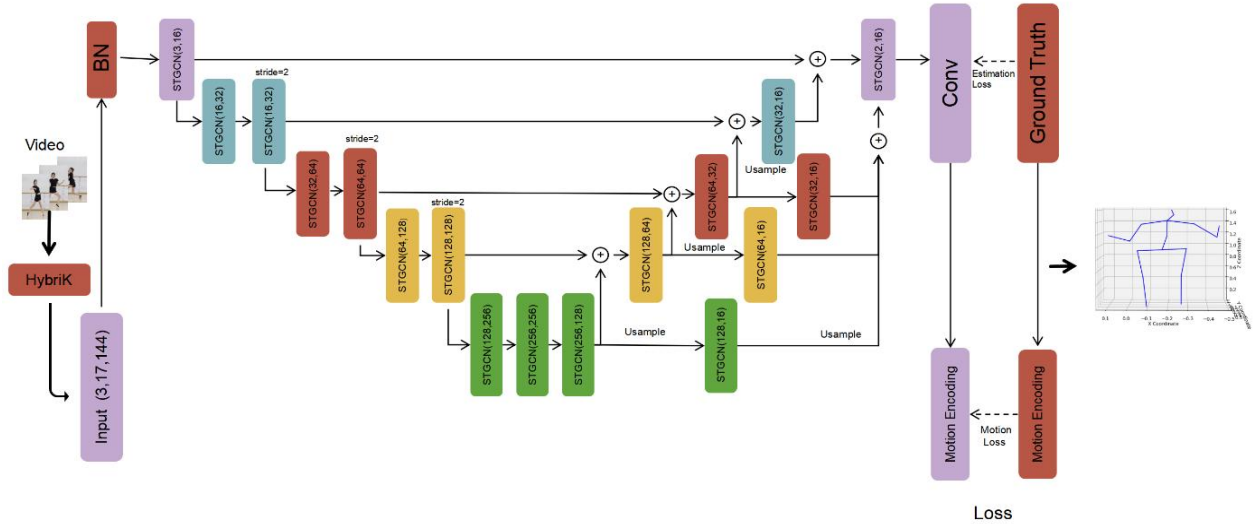

**Figure 1** Network Infrastructure

To determine which 3D human posture key points most closely matched the identified 2D key points, Chen et al.[4]used a large dataset to compare the 2D key points at different views with respect to human posture. Moreno-Noguer[5] In the end, the distance matrix's output provides the essential details of three-dimensional human posture. Li et al.[6] present a method to generate many 3D key points and then extract the most consistent 3D human pose key points through 2D key point mapping. Initially Martinez et al.[7] , it was suggested to finish the generation of 3D skeleton points using a lightweight, straightforward base-line based on 2D skeleton points. Wang et al.[8] used the spatial-temporal map convolution paradigm to create a model shaped like UNet. The long-distance information aggregation is completed by the convolution model of the U-shaped space-time graph[9].In order for the network to have the qualities of a lightweight, straightforward network structure and quick training speed, this technique relies on the maturity of the 2D pose estimation method and performs 3D human pose estimation on this basis; nevertheless, it also requires a sufficient amount of sample data.

## 2.2 Convolution of Spatial-Temporal Graphs

Convolutional neural networks were the primary method utilized in deep learning for early human posture estimation. The first instance will be GCN[10] from ST-GCN[11]. It is used in the area of motion recognition for humans. Each video frame's hu-man skeletal information is collected, and when paired with the time dimension, this information is used to determine the action sequence[12]. To extract time and space details about human movement from data based on graph convolution GCN, ST-GCN combines time graph con-volution with space graph convolution. As seen in Figure 2, the input skeleton space-time sequence is formed by correlating each frame's human skeleton with the time level.

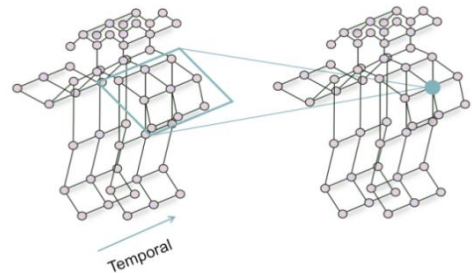

**Figure 2** Skeleton Spatial-Temporal Sequence

## 3 Method

### 3.1 Overview

This section details the architecture and methodology of our 3D-UGCN designed for 3D hu-man pose estimation refinement. Our proposed 3D-UGCN model first outputs the estimated 3D human pose keypoints from the hybrik model, and then 3D-UGCN is used to refine the 3D human pose estimation.

### 3.2 Network Architecture

The direct regression method used in this paper's 3D human posture estimate is based on previous research[13]. Two principles in the HybrIK Model are proposed in the literature[13]: 3D key point estimation and model-based estimation. 3D Key Points Estimation incorporates key points from the human body to create a realistic representation. Both, though, have flaws of their own. Whereas 3D key point estimation is sufficiently precise but may contain certain non-human traits, model-based 3D posture is consistent with human characteristics but not enough accurate. Make the most of both worlds, then. Precise three-dimensional joints are translated into corresponding body part rotations in order to accomplish three-dimensional body mesh reconstruction. As seen in Formula 2, forward kinematics (FK) determines the end key position Q from the initial key point location T and the relative rotation R.

$$q_k = R_k(t_k - t_{pa(k)}) + q_{pa(k)} \#(1)$$

The mathematical procedure of determining the relative rotation R based on the body joint location P is known as inverse kinematics (IK), as Formula 3 illustrates. Simultaneously, a unique hybrid inverse kinematics solution is proposed. via CNN[14] Forecast important 3D areas as you add twist angles, shape characteristics, and rest places. The fundamental static position of the human body is recognized to be the model-based pose. Following the hybridization of these parameters, the relative rotation angle and posture parameters are computed, a mesh is produced, and a three-dimensional human pose that complies with the findings of the human body is recreated.

$$p_k - p_{pa(k)} = R_k(t_k - t_{pa(k)}) \forall 1 \leq k \leq K \#(2)$$

However, the direct regression may cause the problem of missing 3D human poses in some frames. Regarding the UGCN model enumerated in the above mentioned 3D human pose estimation method based on 2D information. The UGCN model takes 2D human information as the input human pose skeleton sequence and employs the UNet network and the ST-GCN network. The Encoder is responsible for gradually extracting the features of the input image and reducing the spatial resolution. The decoder, on the other hand, restores the feature map to the size of the original input image through up-sampling operations and generates the segmentation results step by step.

Secondly, using an input skeleton spatial-temporal sequence for feature extraction, the ST-GCN model builds a skeleton spatial-temporal topological structure diagram. Every frame image's key point is considered a graph node, and two key points that are next to each other create an edge inside the same frame. The network model of HybrIK is shown in Figure 3. A skeletal space graph in time and space is created by connecting the same key points from several frames to generate edges, G=(V,E), where V stands for nodes and E for the edges that connect nodes. The skeleton sequence consists of T frames, $V = \{V_{ti} | t = 1,2,\cdots T; i = 1,2,\cdots N\}$, with N key points in each frame, for all nodes where V is configured to contain the human body position. Two types of E settings exist: the first is the set in space, $E_s = \{v_{ti}v_{tj} | (i,j) \epsilon H\}$, which consists of the sides of the key points connected in the skeleton; the other is the set in time $E_f = \{v_{ti}v_{(t+1)j}\}$, which consists of the motion trajectory of the same node on time change as well as the time connection of the same key point on various frames. It then performs multi-layer graph convolution and time convolution operations on the joint coordinate vector composed of graph nodes and the adjacency matrix composed of graph edges [12].

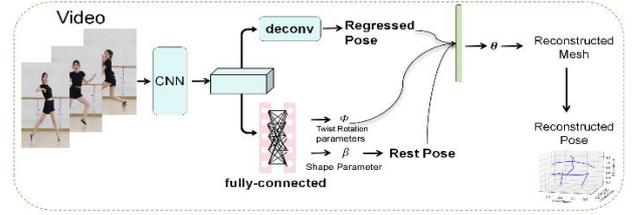

**Figure 3** Network Model of HybrIK

Our proposed 3D-UGCN model uses 3D human body information as an input human posture skeleton sequence constructed into a skeleton spatial-temporal topological map structure for feature extraction. The ST-GCN module is the fundamental unit of the three stages of the UGCN model network, which are down-sampling, up-adoption, and fusion. The input data is first normalized using a batch normalization layer during the down-sampling stage. Next, the number of channels is increased from 3 to 16, 32, 64, 128, and 256 channels by the usage of an ST-GCN module. Finally, the temporal information is extracted using nine ST-GCN modules. Simultaneously, the step size in the module with an even ordinal number is set to 2, increasing the receptive field. There are four ST-GCN modules in the up-sampling phase, each of which is followed by a higher adoption layer. By performing successive up-sampling processes, the temporal resolution can progressively recover and extend to encompass the entire map. The feature maps with varying temporal scales from the second stage are resized to the same size and combined in the fusion step to produce the final embedding. In Figure 4, the network structure diagram is displayed.

## 4 Experimental results and methods

### 4.1 Datasets

The public data collection Human3.6M is used in this work. In the discipline of human posture estimation, it is the largest indoor data set to date. The data set includes three viewing perspectives that are gathered at a rate of fifty hertz (Hz), together with precise motion capture data that represents human poses in three dimensions. These comprised 3.6 million video frames that were recorded by 11 specialists covering around 15 different daily activities. S9 and S11 were the test sets, whereas S1, S5, S6, S7, and S8 were the training sets. To demonstrate the effectiveness of our proposed method, so we evaluate it on the MPI-INF-3DHP[16] dataset as shown in Table I.

**Table I** Comparison results on the MPI-INF-3DHP dataset

|  | HybrIK | UGCN | 3D-UGCN |
|---|---|---|---|
| **MPJPE** | 91.0 | 125.90 | 82.6 |

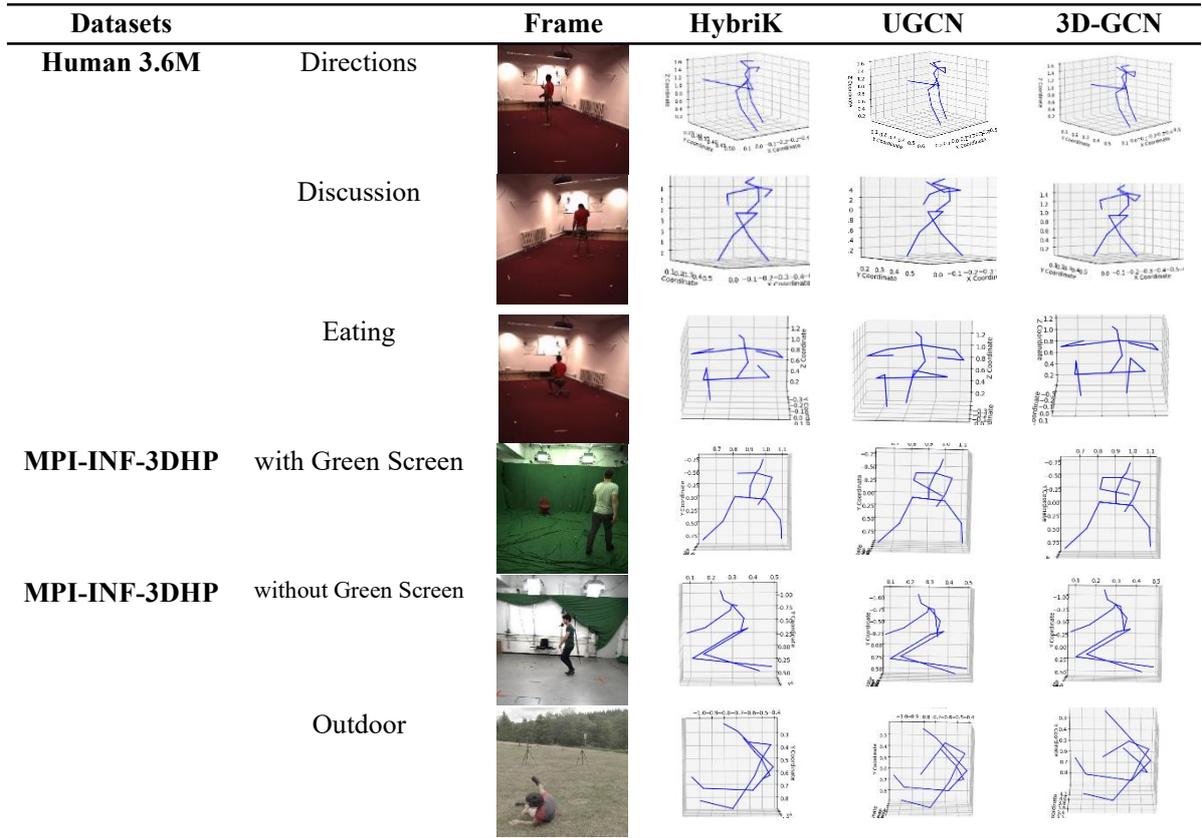

**Figure 4** Comparison chart of experimental results.

### 4.2 Evaluation indicators

In this study, the term "Mean Per Joint Position Error" (MPJPE) refers to the average error of each joint position. It is produced by averaging the difference (single-bit mm) between each node's estimated coordinates and the correct coordinates[17]. Formula 1 looks like this:

$$MPJPE = \frac{1}{T}\frac{1}{N}$$

In addition, the loss function in this paper only calculates the MPJPE between the predicted 3D position and the real position.

### 4.4 Experimental Results

#### 4.4.1 Quantitative Results

The results of our approach on the Human3.6M dataset, which includes approaches based on 2D information boosted to 3D human poses as well as monocular 3D pose estimation methods, are displayed in Table II.

#### 4.4.2 Qualitative Results

In order to show the results more intuitively, we visualize the output of the 3D human pose estimation. As shown in Fig. 4, we present the results of HybriK, UGCN and 3D-UGCN in respectively. When occlusion

$$\sum_{t=1}^{T}\sum_{i=1}^{N} \| \left(J_i^{(t)} - J_{root}^{(t)}\right) - \left(\hat{J}_i^{(t)} - \hat{J}_{root}^{(t)}\right) \| \quad (3)$$

### 4.3 Experimental details

The experimental environment of this paper is Ubuntu 22.04, using NVIDIA RTX 4090 graphics card, CUDA version 11.3, Python version 3.8, batch size set to 256, and training peri-od of 110 rounds. The initial learning rate is 0.01. The dropout rate of each dropout layer is 0.05.

exists, the traditional algorithms tend to have breaks or discontinuities in the critical parts of the skeleton, which seriously affects the quality of the final pose estimation. In contrast, the algorithm proposed in this paper has significant advantages in dealing with the occlusion problem.

## 5  Conclusions

By integrating temporal and geographical variables, the 3D-UGCN network presented in this research successfully raises the accuracy of human posture estimation. The 3D-UGCN network achieves good performance in the experimental results on the Human3.6M dataset. The UGCN network performs well in human posture estimation tasks as evidenced by the noticeably lower MPJPE evaluation index. Still, the network has certain shortcomings, namely its incapacity to effectively manage occlusion and complicated positions. Subsequent investigations may concentrate on strengthening the network's capacity to manage these difficulties and raising the precision of human posture estimate even more.

Table II MPJPE results for 3D pose estimation on the Human3.6M dataset

| MODEL | Direct. | Discuss | Eating | Greet | Photo | Pose | Purch. | Sitting | SittingD | Smoke | Phone | Wait | Walk | WalkD | WalkT | Avg |
|---|---|---|---|---|---|---|---|---|---|---|---|---|---|---|---|---|
| ordinal-pose3d[18] | 48.5 | 54.4 | 54.4 | 52.0 | 59.4 | 65.3 | 49.9 | 52.9 | 65.8 | 71.1 | 56.6 | 52.9 | 60.9 | 44.7 | 47.8 | 56.2 |
| c2f-vol-train[19] | 67.4 | 72.0 | 66.7 | 69.1 | 72.0 | 77.0 | 65.0 | 68.3 | 83.7 | 96.5 | 71.7 | 65.8 | 74.9 | 59.1 | 63.2 | 71.5 |
| 3dpose-demo-iitb[20] | 44.8 | 50.4 | 44.7 | 49.0 | 52.9 | 43.5 | 45.5 | 63.1 | 87.3 | 51.7 | 61.4 | 48.5 | 37.6 | 52.2 | 41.9 | 52.1 |
| Anatomy3D [21] | 41.4 | 43.5 | 40.1 | 42.9 | 46.6 | 51.9 | 41.7 | 42.3 | 53.9 | 60.2 | 45.4 | 41.7 | 46.0 | 31.5 | 32.7 | 44.1 |
| HybriK | 50.2 | 54.7 | 48.1 | 49.2 | 53.5 | 47.6 | 49.7 | 63.6 | 87.8 | 58.1 | 62.9 | 51.9 | 38.3 | 56.0 | 44.4 | 54.4 |
| UGCN (2d) | 57.98 | 68.32 | 75.25 | 65.25 | 77.87 | 63.40 | 68.24 | 87.49 | 119.96 | 77.46 | 95.18 | 68.85 | 90.31 | 69.21 | 67.73 | 76.83 |
| Ours (3d) | 41.16 | 44.30 | 40.88 | 41.82 | 44.94 | 53.53 | 46.04 | 46.36 | 57.57 | 62.74 | 49.80 | 47.53 | 54.8 | 38.44 | 40.34 | 47.34 |

## Acknowledgements

This work is supported by the fund of Beijing Municipal Education Commission 22019821001, the Zhiyuan Science Foundation of BIPT (No.2023014), and Undergraduate Research Training(2024J00012).

## References

[1] Wei G, Yangming Wu, Hong Cui et al. RCVNet: A bird damage identification network for power towers based on fusion of RF images and visual images. Advanced Engineering Informatics, 2023, 57: 102104.
[2] Yean Z, Wei L, Ruoqi Z et al. Dual-channel cascade pose estimation network trained on infrared thermal image and groundtruth annotation for real-time gait measurement. Medical Image Analysis, 2022, 79: 102435.
[3] Wen Wu Y, Yue L, Shuai X, et al. Lightweight multi-person motion capture system in the wild. Sci Sin Inform, 2023, 53(11): 2230–2249.
[4] Ching-Hang C, Deva R. 3d human pose estimation= 2d pose estimation+ matching. Proceedings of the IEEE conference on computer vision and pattern recognition,2017: 7035-7043.
[5] Francesc M.-N.3d human pose estimation from a single image via distance matrix regression. Proceedings of the IEEE conference on computer vision and pattern recognition,2017: 2823-2832.
[6] Li C, Lee G H. Generating multiple hypotheses for 3d human pose estimation with mixture density network. Proceedings of the IEEE/CVF conference on computer vision and pattern recognition,2019: 9887-9895.
[7] Julieta M, Rayat H, Javier R et al. A simple yet effective baseline for 3d human pose estimation. Proceedings of the IEEE international conference on computer vision,2017: 2640-2649.
[8] Jingbo W, Sijie Y, Yuanjun X et al. Motion guided 3d pose estimation from videos. European conference on computer vision, 2020: 764-780.
[9] Olaf R, Philipp F, Thomas B. U-net: Convolutional networks for biomedical image segmentation. Medical image computing and computer-assisted intervention–MICCAI 2015,2015: 234-241.
[10] Thomas N. K, Max W. Semi-supervised classification with graph convolutional networks.arXiv:1609.02907, 2016.
[11] Sijie Y, Yuanjun X, Dahua L. Spatial temporal graph convolutional networks for skeleton-based action recognition. Proceedings of the AAAI conference on artificial intelligence,2018, 32(1).
[12] Tao F, Yanping L, Rui W. Improved human action recognition model based on graph convolutional networks. Electronic Measurement Technology,2023,46(8):59-64.
[13] Jiefeng L, Chao X, Zhicun C et al. Hybrik: A hybrid analytical-neural inverse kinematics solution for 3d human pose and shape estimation. Proceedings of the IEEE/CVF conference on computer vision and pattern recognition,2021: 3383-3393.
[14] Yahui C. Convolutional neural network for sentence classification. Master's thesis. University of Waterloo. 2015
[15] Catalin I; Dragos P; Vlad O et al. Human3. 6m: Large scale datasets and predictive methods for 3d human sensing in natural environments. IEEE transactions on pattern analysis and machine intelligence,2013,36(7): 1325-1339.
[16] Dushyant M, Helge R, Dan C et al. Monocular 3D human pose estimation in the wild using improved cnn supervision. International Conference on 3D Vision (3DV), 2017,506–516.
[17] Faming W, Jianwei L, Sixi C. Overview of Research on 3D Human Pose Estimation. Computer Engineering and Applications,2021,57(10):26-38.
[18] Georgios P, Xiaowei Z, Kostas D. Ordinal depth supervision for 3d human pose estimation. Proceedings of the IEEE conference on computer vision and pattern recognition. 2018: 7307-7316.
[19] Georgios P, Xiaowei Z, Konstantinos G. D, et al. Coarse-to-fine volumetric prediction for single-image 3D human pose. Proceedings of the IEEE conference on computer vision and pattern recognition. 2017: 7025-7034.
[20] Rishabh D, Anurag M, Uday K et al. Learning 3d human pose from structure and motion. Proceedings of the European conference on computer vision. 2018: 668-683.
[21] Tianlang C, Chen F, Xiaohui S et al. Anatomy-aware 3d human pose estimation with bone-based pose decomposition. IEEE Transactions on Circuits and Systems for Video Technology.2021, 32(1): 198-290.